\documentclass{ifacconf}
\usepackage{amsmath}
\DeclareMathOperator*{\argmin}{arg\,min}    
\newcommand{\norm}[1]{\left\lVert#1\right\rVert} 
\usepackage[makeroom]{cancel}   
\usepackage{amssymb}
\usepackage{graphicx}      
\usepackage{natbib}        
\usepackage{xcolor}
\usepackage{soul}
\usepackage{url}
\begin{document}
\begin{frontmatter}
\title{Distributed IDA-PBC for a Class of Nonholonomic Mechanical Systems} 
\author[First]{A. Tsolakis} 
\author[Second]{T. Keviczky} 
\address[First]{Cognitive Robotics, Delft University of Technology, 2628 CD, Delft, The Netherlands (e-mail: a.tsolakis@tudelft.nl)}
\address[Second]{Delft Center for Systems and Control, Delft University of Technology, 2628 CD, Delft, The Netherlands (t.keviczky@tudelft.nl)}
\begin{abstract}                
Nonholonomic mechanical systems encompass a large class of practically interesting robotic structures, such as wheeled mobile robots, space manipulators, and multi-fingered robot hands. However, few results exist on the cooperative control of such systems in a generic, distributed approach. In this work we extend a recently developed distributed Interconnection and Damping Assignment Passivity-Based Control (IDA-PBC) method to such systems. More specifically, relying on port-Hamiltonian system modelling for networks of mechanical systems, we propose a full-state stabilization control law for a class of nonholonomic systems within the framework of distributed IDA-PBC. This enables the cooperative control of heterogeneous, underactuated and nonholonomic systems with a unified control law. This control law primarily relies on the notion of Passive Configuration Decomposition (PCD) and a novel, non-smooth desired potential energy function proposed here. A low-level collision avoidance protocol is also implemented in order to achieve dynamic inter-agent collision avoidance, enhancing the practical relevance of this work. Theoretical results are tested in different simulation scenarios in order to highlight the applicability of the derived method.
\end{abstract}
\end{frontmatter}
\section{Introduction}\label{ss:intro}
An increasing demand in multi-agent systems has been spurred by the benefits obtained when a single complex system is transformed to an equivalent set of multiple yet simpler systems. With a distributed control architecture, lower level components operate on local information in an appropriate manner to accomplish global goals. This decomposition of a complex system into simpler units and their distributed control entails great advantages among which, decreased operational cost, robustness to failure, strong adaptivity and system scalability \citep{Cao2012}. Distributed control of mechanical systems can be used in numerous applications such as collaborative transportation, exploration of unknown or dangerous terrains, large scale sensing and area monitoring, collaborative construction and vehicle platoons or spacecraft constellations.

The dynamics of mechanical systems are highly nonlinear. Feedback stabilization of nonlinear systems has occupied a central role in the literature of nonlinear systems. A class of nonlinear control methods known as passivity-based control has been proven to be especially suitable. These methods rely on the fundamental property of passivity  which is instrumental for deriving stabilizing control laws. The nonlinear system can be controlled by shaping its closed loop energy while respecting the original dynamics, an intrinsically less conservative method which provides higher performance, cost-effective controllers. The most general method that combines total energy shaping and damping injection is Interconnection- and Damping Assignment Passivity-Based Control (IDA-PBC) \citep{ortega2002}. This method assumes that the system admits to Hamiltonian dynamics, which are inherently passive. Recently, a distributed IDA-PBC scheme for fully- and underactuated mechanical systems was developed in \citet{valk2018}. While this method shows a great potential for many robotic applications, it is not yet applicable to the wide class of nonholonomic mechanical systems. The evolution of such systems is dictated not only by the equations of motion but also an additional set of non-integrable differential equations known as nonholonomic constraints. These constraints introduce a coupling among the system's generalized velocities complicating the analysis. Control of nonholonomic systems often relied on appropriate coordinate transformations \citep{astolfi1996, fujimoto2012}, while similar approaches are also used in the cooperative control of such systems \citep{dong2008,qu2008,du2016}. While these are powerful techniques, the majority of them relies on specific control forms (e.g., normal form, power form, chained form, etc.), that can be attained typically only by using feedback linearization (i.e., cancellation of nonlinear dynamics) and state transformation as stated in \cite{lee2016}. Moreover, most of the aforementioned cooperative control methods focus on the kinematic control of homogeneous teams of agents rather than the full dynamic control of heterogeneous teams in which nonholonomic (e.g., wheeled mobile robots) and underactuated (e.g., quad-rotors) agents need to cooperate together.

In this paper we propose a method with which we can extend the results of \citet{valk2018} to a class of nonholonomic mechanical systems. This method relies on the notion of Passive Configuration Decomposition (PCD) from \citet{lee2016} and a novel desired potential function proposed here with which we can achieve smooth stabilization in the constrained space. In order to enhance the practical relevance of this work we also implement a simple collision avoidance protocol based on the method of Artificial Potential Fields (APF) and some of its extensions. We show the efficacy of  the theoretical results via different simulation scenarios. 

In Sections \ref{ss:nonholsystems} and \ref{ss:nonholcontrol} we review the Hamiltonian formulation and control with IDA-PBC for nonholonomic systems which establishes the foundations of this work. Section \ref{ss:nonholstabilization} presents the main results which are two-fold: the adaptation of PCD to Hamiltonian systems, and the proposal of a novel desired potential function, respectively. In Section \ref{ss:results}, we illustrate simulation results along with a critical discussion. We conclude this work with a few remarks and recommendations for future work in Section \ref{ss:conclusion}.

\section{Hamiltonian Formulation of Nonholonomic Mechanical Systems}\label{ss:nonholsystems}
In this section we review the Hamiltonian formulation of nonholonomic mechanical systems as derived in \cite{vanderschaft1994} for completeness. We are interested in mechanical systems which are subjected to nonholonomic constraints in Pfaffian form. The frictionless dynamics of a nonholonomic, mechanical system with generalized coordinates $\boldsymbol{q} \in \mathbb{R}^n$, generalized momenta $\boldsymbol{p} = \boldsymbol{M}(\boldsymbol{q}) \boldsymbol{\dot{q}} \in \mathbb{R}^n$, constraint forces $\boldsymbol{\lambda} \in \mathbb{R}^k$, input $\boldsymbol{\tau} \in \mathbb{R}^m$  and conjugate output $\boldsymbol{y} \in \mathbb{R}^m$ are expressed as:
\begin{equation}\label{eq:ham_exp_constr_eom}
    \left[\begin{array}{c}
    {\boldsymbol{\dot{q}}} \\
    {\boldsymbol{\dot{p}}}
    \end{array}\right] 
    =
    \left[\begin{array}{cc}
    {\boldsymbol{0}_{n}} & {\boldsymbol{I}_{n}} \\
    {-\boldsymbol{I}_{n}} & {\boldsymbol{0}_{n}}
    \end{array}\right]
    \left[\begin{array}{c}
    {\frac{\partial H}{\partial \boldsymbol{q}}(\boldsymbol{q}, \boldsymbol{p})} \\
    {\frac{\partial H}{\partial \boldsymbol{p}}(\boldsymbol{q}, \boldsymbol{p})}
    \end{array}\right]
    +
    \left[\begin{array}{c}
    {\boldsymbol{0}_{n \times k}} \\
    {\boldsymbol{A}(\boldsymbol{q})}
    \end{array}\right] \boldsymbol{\lambda}
    +
    \left[\begin{array}{c}
    {\boldsymbol{0}_{n \times m}} \\
    {\boldsymbol{F}(\boldsymbol{q})}
    \end{array}\right]
    \boldsymbol{\tau}
\end{equation}
\begin{equation}\label{eq:ham_exp_constr_out}
    \boldsymbol{y} 
    =
    \boldsymbol{F}^{\top}(\boldsymbol{q}) \frac{\partial H}{\partial \boldsymbol{p}}(\boldsymbol{q}, \boldsymbol{p})
\end{equation}
\begin{equation}\label{eq:ham_exp_constr_con}
    \boldsymbol{0} 
    =
    \boldsymbol{A}^{\top}(\boldsymbol{q}) \frac{\partial H}{\partial \boldsymbol{p}}(\boldsymbol{q}, \boldsymbol{p})
\end{equation}
\begin{equation}\label{eq:ham_exp_constr_ham}
    H
    = 
    \frac{1}{2}\boldsymbol{p}^{\top} \boldsymbol{M}^{-1} \boldsymbol{p} + V(\boldsymbol{q})
\end{equation}
where $\boldsymbol{M}(\boldsymbol{q})=\boldsymbol{M}^{\top}(\boldsymbol{q})>\boldsymbol{0}_n$ is the generalized mass matrix, $\boldsymbol{A}(\boldsymbol{q}) \in \mathbb{R}^{n \times k}$ the constraint matrix with $rank(\boldsymbol{A})=k < n$ and $\boldsymbol{F}(\boldsymbol{q}) \in \mathbb{R}^{n \times m}$ the input matrix with $rank(\boldsymbol{F})=m < n$. The Hamiltonian $H(\boldsymbol{q},\boldsymbol{p})$ is the system's mechanical energy given as the sum of kinetic energy $\frac{1}{2}\boldsymbol{p}^{\top} \boldsymbol{M}^{-1} \boldsymbol{p}$ and potential energy $V(\boldsymbol{q}) \in \mathbb{R}$. The system's state is the pair $(\boldsymbol{q},\boldsymbol{p}) \in \mathcal{X}$. The constraint equation \eqref{eq:ham_exp_constr_con} appears explicitly in the system description complicating the analysis and control of these systems. An efficient way to work with such systems is to express the equations of motion on the constrained space. Since $rank(\boldsymbol{A}(\boldsymbol{q}))=k$, there exists locally a smooth matrix $\boldsymbol{S}(\boldsymbol{q}) \in \mathbb{R}^{n \times (n-k)}$ of rank $n-k$ such that:
\begin{equation}\label{eq:annihilator_s}
    \boldsymbol{A}^{\top}(\boldsymbol{q})\boldsymbol{S}(\boldsymbol{q}) = \boldsymbol{0}_{k \times (n-k)}
\end{equation}
Now define $\boldsymbol{\tilde{p}} = \boldsymbol{T}(\boldsymbol{q})\boldsymbol{p}$ with the transformation matrix chosen as in \citet{muralidharan2009}:
\begin{equation}\label{eq:transformation_matrix}
    \boldsymbol{T}({\boldsymbol{q}}) = 
    \left[\begin{array}{c}
    {\boldsymbol{S}^{\top}(\boldsymbol{q})} \\
    {\boldsymbol{A}^{\top}(\boldsymbol{q}) \boldsymbol{M}^{-1}(\boldsymbol{q})}
    \end{array}\right] \in \mathbb{R}^{n \times n}
\end{equation}
Partitioning the generalised momenta as $\boldsymbol{\tilde{p}} = \begin{pmatrix} \boldsymbol{\tilde{p}^1} \\ \boldsymbol{\tilde{p}^2} \end{pmatrix}$ yields:
\begin{equation}\label{eq:momenta_partition}
    \boldsymbol{\tilde{p}^1} = \boldsymbol{S}^{\top}(\boldsymbol{q})\boldsymbol{p} \in \mathbb{R}^{n-k}, \quad \boldsymbol{\tilde{p}^2} = \boldsymbol{A}^{\top}(\boldsymbol{q}) \boldsymbol{M}^{-1}(\boldsymbol{q})  \boldsymbol{p} \in \mathbb{R}^{k}
\end{equation}
Notice that $\boldsymbol{\tilde{p}^2} = \boldsymbol{0}$ over the constrained manifold because of \eqref{eq:ham_exp_constr_con}. Moreover the introduction of the annihilator removes the constraint forces from \eqref{eq:ham_exp_constr_eom}. For the sake of notation we denote here $\boldsymbol{\tilde{p}^1} \triangleq \boldsymbol{\tilde{p}} \in \mathbb{R}^{n-k}$. Thus, equations \eqref{eq:ham_exp_constr_eom}-\eqref{eq:ham_exp_constr_ham} can be equivalently written as:
\begin{equation}\label{eq:ham_imp_constr_eom}
    \left[\begin{array}{c}
    {\boldsymbol{\dot{q}}} \\
    {\boldsymbol{\dot{\tilde{p}}}}
    \end{array}\right] 
    =
    \left[\begin{array}{cc}
    {\boldsymbol{0}}_{n} & {\boldsymbol{S}(\boldsymbol{q})} \\
    {-\boldsymbol{S}^{\top}(\boldsymbol{q})} & {\boldsymbol{Y}(\boldsymbol{q},\boldsymbol{\tilde{p}})}
    \end{array}\right]
    \left[\begin{array}{c}
    {\frac{\partial \tilde{H}}{\partial \boldsymbol{q}}(\boldsymbol{q}, \boldsymbol{\tilde{p}})} \\
    {\frac{\partial \tilde{H}}{\partial \boldsymbol{\tilde{p}}}(\boldsymbol{q}, \boldsymbol{\tilde{p}})}
    \end{array}\right]
    +
    \left[\begin{array}{c}
    {\boldsymbol{0}_{n \times m}} \\
    {\boldsymbol{\tilde{F}}(\boldsymbol{q})}
    \end{array}\right]
    \boldsymbol{\tau}
\end{equation}
\begin{equation}\label{eq:ham_imp_constr_out}
    \boldsymbol{\tilde{y}}
    =
    \boldsymbol{\tilde{F}}^{\top}(\boldsymbol{q}) \frac{\partial \tilde{H}}{\partial \boldsymbol{\tilde{p}}}(\boldsymbol{q}, \boldsymbol{\tilde{p}})
\end{equation}
\begin{equation}\label{eq:ham_imp_constr_ham}
    \tilde{H}(\boldsymbol{q}, \boldsymbol{\tilde{p}})
    =
    \frac{1}{2} \boldsymbol{\tilde{p}}^{\top} \boldsymbol{\tilde{M}}^{-1}(\boldsymbol{q}) \boldsymbol{\tilde{p}}+V(\boldsymbol{q})
\end{equation}
where $\boldsymbol{\tilde{y}}(\boldsymbol{q}, \boldsymbol{\tilde{p}}) \in \mathbb{R}^m$ is the transformed output, $\tilde{H}(\boldsymbol{q}, \boldsymbol{\tilde{p}}) \in \mathbb{R}$ the transformed Hamiltonian, $\boldsymbol{\tilde{M}}(\boldsymbol{q}) = \boldsymbol{S}^{\top} \boldsymbol{M} \boldsymbol{S} > \boldsymbol{0}_{n-k}$ the symmetric transformed generalized mass matrix, $\boldsymbol{\tilde{F}}(\boldsymbol{q}) =  \boldsymbol{S}^{\top} \boldsymbol{F} \in \mathbb{R}^{(n-k)  \times m}$, the transformed input matrix and $\boldsymbol{Y} = \left( -\boldsymbol{p}^T[S_i,S_j](\boldsymbol{q}) \right) _{i,j=1,...,n-k} \in \mathbb{R}^{(n-k) \times (n-k)} $ is a skew-symmetric matrix that arises from the existence of constraints with $[S_i,S_j] $ denoting the Lie bracket. More elaborate expressions for these components can be found in \citet{tsolakis2021}. The new system is expressed in the new set of coordinates $(\boldsymbol{q},\boldsymbol{\tilde{p}}) \in \mathbb{R}^{2n-k}$ and evolves on the constrained manifold  $\mathcal{X}_c$. It is described by a set of $2n-k$ nonlinear, input-affine ODEs with an $m$-dimensional input $\boldsymbol{\tau}$ and a set of $2n-k$ initial conditions $\boldsymbol{x_0} = (\boldsymbol{q_0}^{\top},\boldsymbol{\tilde{p}_0}^{\top})^{\top} \in \mathbb{R}^{2n-k}$ which can be derived using the transformation matrix defined in \eqref{eq:transformation_matrix}.

\section{IDA-PBC for a class of Nonholonomic Mechanical Systems}\label{ss:nonholcontrol}
In this section we want to apply the classical IDA-PBC method of \citet{ortega2002} to the nonholonomic systems described by equations \eqref{eq:ham_imp_constr_eom}-\eqref{eq:ham_imp_constr_ham}. The first general adaptation to nonholonomic systems is found in \citet{blankenstein2002} considering nonholonomic systems that may be underactuated in the constrained space. Assuming that the nonholonomic systems we are interested in are \textit{fully-actuated in the constrained space}, the desired dynamics can take the following form as in \citet{muralidharan2009}:
\begin{equation}\label{eq:ham_desired_eom}
    \left[\begin{array}{c}
    {\boldsymbol{\dot{q}}} \\
    {\boldsymbol{\dot{\tilde{p}}}}
    \end{array}\right] 
    =
    \left[\begin{array}{cc}
    {\boldsymbol{0}_{n}} & {\boldsymbol{S}\boldsymbol{\tilde{M}}^{-1}\boldsymbol{M_d}} \\
    {-\boldsymbol{M_d}\boldsymbol{\tilde{M}}^{-1}\boldsymbol{S}^{\top}} & {\boldsymbol{J}-\boldsymbol{\tilde{F}}\boldsymbol{K_v}\boldsymbol{\tilde{F}}^{\top}}
    \end{array}\right]
    \left[\begin{array}{c}
    {\frac{\partial H_d}{\partial \boldsymbol{q}}(\boldsymbol{q}, \boldsymbol{\tilde{p}})} \\
    {\frac{\partial H_d}{\partial \boldsymbol{\tilde{p}}}(\boldsymbol{q}, \boldsymbol{\tilde{p}})}
    \end{array}\right]
\end{equation}
\begin{equation}\label{eq:ham_desired_out}
    \boldsymbol{y_d} 
    =
    \boldsymbol{\tilde{F}}^{\top}(\boldsymbol{q}) \frac{\partial H_d}{\partial \boldsymbol{\tilde{p}}}(\boldsymbol{q}, \boldsymbol{\tilde{p}})
\end{equation}
\begin{equation}\label{eq:ham_desired_ham}
    H_d(\boldsymbol{q}, \boldsymbol{\tilde{p}}) 
    =
    \frac{1}{2} \boldsymbol{\tilde{p}}^{\top} \boldsymbol{M_d}^{-1}(\boldsymbol{q}) \boldsymbol{\tilde{p}}+V_d(\boldsymbol{q})
\end{equation}
where $\boldsymbol{M_d} \in \mathbb{R}^{(n-k) \times (n-k)}$ is the \textit{desired mass matrix} which shapes the kinetic energy, and $V_d \in \mathbb{R}$ is the \textit{desired potential energy} which shapes the potential energy. The desired potential energy $V_d$ aims to make the system evolve towards a goal configuration denoted as $\boldsymbol{q^*}$ thus having the property:
\begin{equation}\label{eq:goal_qstar}
    \boldsymbol{q^*} = \argmin_{\boldsymbol{q}} V_d(\boldsymbol{q})
\end{equation}
Kinetic energy shaping aims to solve the matching problem and in addition shapes the transient response. Matrix $\boldsymbol{J}=-\boldsymbol{J}^{\top} \in \mathbb{R}^{(n-k) \times (n-k)}$ is the skew-symmetric, \textit{gyroscopic force matrix} which aids in the solution of the matching problem as well, by creating one extra degree of freedom in the matching conditions as explained later. The \textit{damping matrix} denoted as $\boldsymbol{K_v} = \boldsymbol{K_v}^{\top} > \boldsymbol{0}_{m}$ induces dissipation to the closed-loop system for asymptotic convergence. This matrix is free to choose as it does not appear in the so-called matching conditions presented next. 

IDA-PBC aims to find a control input $\boldsymbol{\tau} \in \mathbb{R}^m$ that transforms the open-loop plant \eqref{eq:ham_imp_constr_eom}-\eqref{eq:ham_imp_constr_ham} to the desired, closed-loop dynamics \eqref{eq:ham_desired_eom}-\eqref{eq:ham_desired_ham}. This is known as the \textit{matching problem} since we need to match the controlled system with the desired dynamics. In order to solve the problem we begin with equating the open-loop dynamics \eqref{eq:ham_imp_constr_eom} with control input $\boldsymbol{\tau} \in \mathbb{R}^m$ to the closed-loop dynamics \eqref{eq:ham_desired_eom}. Following the classical approach as in \citet{ortega2002}, this yields the control law:
\tiny
\begin{equation}\label{eq:feedback_law}
\begin{gathered}
\boldsymbol{\tau}=\left(\boldsymbol{\tilde{F}}^{\top} \boldsymbol{\tilde{F}}\right)^{-1} \boldsymbol{\tilde{F}}^{\top}\left(\boldsymbol{S}^{\top} \frac{\partial \tilde{H}}{\partial \boldsymbol{q}}-\boldsymbol{M_{d}} \boldsymbol{\tilde{M}}^{-1} \boldsymbol{S}^{\top} \frac{\partial H_{d}}{\partial \boldsymbol{q}}-\boldsymbol{Y}\frac{\partial \tilde{H}}{\partial \boldsymbol{\tilde{p}}}+\boldsymbol{J} \frac{\partial H_{d}}{\partial \boldsymbol{\tilde{p}}}\right)\\
-\boldsymbol{K_{v}} \boldsymbol{\tilde{F}}^{\top} \frac{\partial H_{d}}{\partial \boldsymbol{\tilde{p}}}
\end{gathered}
\end{equation}
\normalsize
and the kinetic and potential matching conditions:
\tiny
\begin{equation}\label{eq:mathcing_cond_kin}
\begin{gathered}
\boldsymbol{\tilde{F}}^{\perp}\left(\boldsymbol{S}^{\top} \frac{\partial \boldsymbol{\tilde{p}}^{\top} \boldsymbol{\tilde{M}}^{-1} \boldsymbol{\tilde{p}}}{\partial \boldsymbol{q}}-\boldsymbol{M_{d}} \boldsymbol{\tilde{M}}^{-1} \boldsymbol{S}^{\top} \frac{\partial \boldsymbol{\tilde{p}}^{\top} \boldsymbol{M_{d}}^{-1} \boldsymbol{\tilde{p}}}{\partial \boldsymbol{q}}-2\boldsymbol{Y}\boldsymbol{\tilde{M}}^{-1}\boldsymbol{\tilde{p}} +2\boldsymbol{J} \boldsymbol{M_{d}}^{-1} \boldsymbol{\tilde{p}}\right)\\ =\boldsymbol{0}_{n-k-m}
\end{gathered}
\end{equation}
\normalsize
\begin{equation}\label{eq:mathcing_cond_pot}
\boldsymbol{\tilde{F}}^{\perp}\left(\frac{\partial V}{\partial \boldsymbol{q}}-\boldsymbol{M_{d}} \boldsymbol{\tilde{M}}^{-1} \boldsymbol{S}^{\top} \frac{\partial V_{d}}{\partial \boldsymbol{q}}\right)=\boldsymbol{0}_{n-k-m}
\end{equation}
where we denote as $\boldsymbol{\tilde{F}^{\bot}} \in \mathbb{R}^{(n-k-m) \times (n-k)}$ the left annihilator of $\boldsymbol{\tilde{F}}$ such that $\boldsymbol{\tilde{F}^{\bot}}\boldsymbol{\tilde{F}} = \boldsymbol{0}_{(n-k-m)\times m}$. The matching conditions \eqref{eq:mathcing_cond_kin} and \eqref{eq:mathcing_cond_pot}  ensure that the control actions are feasible in case the system is underactuated. In the case of holonomic systems, the system description and control input reduce to the original form as in \citet{ortega2002}. For holonomic systems, we can choose suitable $\boldsymbol{M_d}$ and $\boldsymbol{J}$ so that the PDEs \eqref{eq:mathcing_cond_kin}-\eqref{eq:mathcing_cond_pot} are satisfied, $\boldsymbol{K_v}$ to inject damping (and thus asymptotic stabilization) and a smooth desired potential $V_d$ with which we can stabilize the system at an arbitrary desired equilibrium given in \eqref{eq:goal_qstar}. However, that is not the case for nonholonomic systems as Brockett's necessary conditions suggests \citep{brockett1983}. Due to the existence of nonholonomic constraints, the system will be stabilized at the largest invariant set:
\begin{equation}\label{eq:invariant_set}
\mathbf{\Omega}_{inv}=\left\{(\boldsymbol{q}, \boldsymbol{0}) \in \mathcal{X} \Big| \boldsymbol{S}^{\top}(\boldsymbol{q}) \frac{\partial V_d}{\partial \boldsymbol{q}}(\boldsymbol{q})=\boldsymbol{0}\right\}
\end{equation}
In the next section we propose a method to tackle this problem.
\section{Stabilization of a Class of Nonholonomic Systems}\label{ss:nonholstabilization}
In this section we propose a method with which we can use the nonholonomic IDA-PBC control law \eqref{eq:feedback_law} derived in the previous section so that the system can be successfully stabilized at the desired equilibrium. The proposed method consists of two parts. The first part is the adaptation of PCD to port-Hamiltonian systems. The PCD introduced in \citet{lee2016}, was applied to the open-loop Lagrangian dynamics of nonholonomic systems. In our case, we apply this method to the closed-loop, Hamiltonian dynamics \eqref{eq:ham_desired_eom}-\eqref{eq:ham_desired_ham} so that we can use the already derived control law \eqref{eq:feedback_law}. There are two main reasons to extend this result in the framework of IDA-PBC. First of all, since IDA-PBC has been proven a favourable approach for underactuated systems, an extension of PCD to port-Hamiltonian systems may allow the development of stabilizing control laws for systems that are both nonholonomic and underactuated such as the Mobile Inverted Pendulum studied in \citet{muralidharan2009}. Moreover, this result is instrumental for the extension of the distributed IDA-PBC method developed originally in \citet{valk2018} to the practically relevant class of nonholonomic systems.  This will allow for distributed cooperative control of a team of heterogeneous systems which may consist of holonomic/nonholonomic, fully-actuated/underactuated mechanical systems thus enhancing the scope of application.  After applying PCD to \eqref{eq:ham_desired_eom}-\eqref{eq:ham_desired_ham}, the second part of the proposed method is a novel choice of the desired potential function $V_d$ that relies on the aforementioned decomposition. More specifically, based on the insight that some of the configuration variables are free from the nonholonomic constraints, we can use the latter to drive the system to the desired equilibrium $\boldsymbol{q^*}$ despite the presence of these constraints. Of course, in order to stabilize the system in the full state space our approach leads to a non-smooth feedback law thus not contradicting with Brockett's necessary condition. 
\subsection{Applying PCD to port-Hamiltonian Systems}\label{ss:PCD_ham}
We are interested in the class of nonholonomic mechanical systems described by \eqref{eq:ham_desired_eom}-\eqref{eq:ham_desired_ham} for which the following assumptions are made \citep{lee2016}:
\vspace{-1mm}
\begin{enumerate}\setlength\itemsep{-0.0em}
    \item The system's configuration space $\mathcal{Q}$ can be endowed with the product structure such that $\mathcal{Q} = \mathcal{S} \times \mathcal{R}$
    with $\boldsymbol{q} =  \begin{pmatrix} \boldsymbol{s}^{\top} & \boldsymbol{r}^{\top} \end{pmatrix}^{\top} ,\ \boldsymbol{s} \in \mathbb{R}^{n-p},\  \boldsymbol{r} \in \mathbb{R}^{p}$.
    \item The constraint matrix of the nonholonomic Pfaffian constraint \eqref{eq:ham_exp_constr_con} is also a function of only $\boldsymbol{r} \in \mathcal{R}$ and the constraint acts only on $\boldsymbol{s} \in \mathcal{S}$:
    \begin{equation}\label{eq:constraint_on_s}
        \boldsymbol{A}^{\top}(\boldsymbol{q}) \boldsymbol{\dot{q}} = 
        \begin{bmatrix}
            \boldsymbol{A_s}^{\top}(\boldsymbol{r})&\boldsymbol{0}_{k \times p}
        \end{bmatrix} \boldsymbol{\dot{q}} = \boldsymbol{A_s}^{\top}(\boldsymbol{r}) \boldsymbol{\dot{s}} = \boldsymbol{0}_{k}
    \end{equation}
    with $\boldsymbol{A_s}(\boldsymbol{r}) \in \mathbb{R}^{(n-p) \times k}$ being full row rank.
    \item  Its inertia matrix is a function of only $\boldsymbol{r} \in \mathcal{R}$, that is, $\boldsymbol{M}(\boldsymbol{q})=\boldsymbol{M}(\boldsymbol{r})$.
\end{enumerate}
The aforementioned properties may seem restrictive but in fact encompass many practically important and interesting systems with some examples listed in \citet{lee2016}. With this structure, the unconstrained distribution $\mathcal{D}_s(\boldsymbol{r}) \in \mathbb{R}^{(n-p) \times (n-p-k)}$ is defined on $\mathcal{S}$ such that:
\begin{equation}\label{eq:distribution_ds}
    \mathcal{D}_s(\boldsymbol{r})^{\top} \boldsymbol{A_s}(\boldsymbol{r}) = \boldsymbol{0}_{(n-p-k) \times k}
\end{equation} 
Since $\boldsymbol{A}(\boldsymbol{r})$ is regular and smooth, so is $\mathcal{D}_s$ with $\mathrm{rank} (\mathcal{D}_s) = n-p-k,\ \forall \boldsymbol{r} \in \mathcal{R}$. We can then partition the mass matrix $\boldsymbol{M}(\boldsymbol{r})$ such that the transformed mass matrix becomes:
\begin{equation}\label{eq:transformed_mass_decomposed}
    \boldsymbol{\tilde{M}}(\boldsymbol{r}) = 
    \boldsymbol{S}(\boldsymbol{r})^{\top} \boldsymbol{M}(\boldsymbol{r}) \boldsymbol{S}(\boldsymbol{r})
    =
    \begin{bmatrix}
    \mathcal{D}_s^{\top}\boldsymbol{M_s}\mathcal{D}_s &  \mathcal{D}_s^{\top}\boldsymbol{M_{sr}} \\
    \boldsymbol{M_{sr}}^{\top}\mathcal{D}_s & \boldsymbol{M_r}
    \end{bmatrix}
\end{equation}
In order to avoid acceleration couplings via the inertia matrix between the $s$-dynamics and the $r$-dynamics which is usually not cancellable, we follow another assumption from \citet{lee2016}: 
\begin{equation}\label{eq:mass_assumption}
    \boldsymbol{M_{sr}^{\top}(\boldsymbol{r})} \mathcal{D}_s(\boldsymbol{r}) = \boldsymbol{0}_{p\times(n-p-k)},\ \forall \boldsymbol{r} \in \mathbb{R}^p
\end{equation}
Thus $ \boldsymbol{\tilde{M}}(\boldsymbol{r})$ becomes block-diagonal leading to decoupling of $\boldsymbol{s}$ and $\boldsymbol{r}$ via the inertia matrix. We can decompose the generalized momenta as $\boldsymbol{\tilde{p}} =  \begin{pmatrix} \boldsymbol{\tilde{p}_s}^{\top} & \boldsymbol{p_r}^{\top} \end{pmatrix}^{\top}$ and choose a block diagonal desired mass matrix $\boldsymbol{M_d}(\boldsymbol{r})$ as:
\begin{equation}\label{eq:desired_mass_decomposed}
    \boldsymbol{M_d}(\boldsymbol{r}) = 
    \begin{bmatrix}
    \boldsymbol{M_{ds}} & \boldsymbol{0} \\
    \boldsymbol{0} & \boldsymbol{M_{dr}}(\boldsymbol{r})
    \end{bmatrix}
\end{equation}
with $\boldsymbol{M_{ds}}$ independent from $\boldsymbol{r}$. The Hamiltonian in \eqref{eq:ham_desired_ham} can be decomposed as:
\small
\begin{equation}\label{eq:hamiltonian_decomposed}
    H_d = 
    \underbrace{\frac{1}{2}\boldsymbol{\tilde{p}_s}^{\top}\boldsymbol{M_{ds}}(\boldsymbol{r}) \boldsymbol{\tilde{p}_s} + V_{ds}(\boldsymbol{s})}_{H_{ds}} + 
    \underbrace{\frac{1}{2}\boldsymbol{p_r}^{\top}\boldsymbol{M_{dr}}\boldsymbol{p_r} + V_{dr}(\boldsymbol{r})}_{H_{dr}} 
\end{equation}
\normalsize
With proper block-diagonal choices for matrices $\boldsymbol{J}$ and $\boldsymbol{K_v}$, the closed-loop system \eqref{eq:ham_desired_eom}-\eqref{eq:ham_desired_ham} can be decomposed to two Hamiltonian systems:
\tiny
\begin{equation}\label{eq:ham_desired_s_eom}
    \left[\begin{array}{c}
    {\boldsymbol{\dot{s}}} \\
    {\boldsymbol{\dot{\tilde{p}}_s}}
    \end{array}\right] 
    =
    \left[\begin{array}{cc}
    \boldsymbol{0}_{n-p} &  \mathcal{D}_s (\mathcal{D}_s^{\top} \boldsymbol{M_s} \mathcal{D}_s)^{-1} \boldsymbol{M_{ds}}
    \\
     -\boldsymbol{M_{ds}} (\mathcal{D}_s^{\top} \boldsymbol{M_s} \mathcal{D}_s)^{-1} \mathcal{D}_s^{\top}  & \boldsymbol{J_s} - \boldsymbol{\tilde{F}_s} \boldsymbol{K_{vs}} \boldsymbol{\tilde{F}_s}^{\top}
    \end{array}\right]
    \left[\begin{array}{c}
    {\frac{\partial H_{ds}}{\partial \boldsymbol{s}}}(\boldsymbol{s}) \\
    {\frac{\partial H_{ds}}{\partial \boldsymbol{\tilde{p}_s}}}(\boldsymbol{\tilde{p}_s})
    \end{array}\right]
\end{equation}
\normalsize
\begin{equation}\label{eq:ham_desired_s_out}
    \boldsymbol{y_{ds}} 
    =
    \boldsymbol{\tilde{F}_s}^{\top} \frac{\partial H_{ds}}{\partial \boldsymbol{\tilde{p}_s}}
\end{equation}
\begin{equation}\label{eq:ham_desired_s_ham}
    H_{ds}
    =
    \frac{1}{2} \boldsymbol{\tilde{p}_s}^{\top} \boldsymbol{M_{ds}}^{-1} \boldsymbol{\tilde{p}_s}+V_{ds}(\boldsymbol{s})
\end{equation}
\tiny
\begin{equation}\label{eq:ham_desired_r_eom}
    \left[\begin{array}{c}
    {\boldsymbol{\dot{r}}} \\
    {\boldsymbol{\dot{p}_r}}
    \end{array}\right] 
    =
    \left[\begin{array}{cc}
    \boldsymbol{0}_{p} &  \boldsymbol{M_r}^{-1} \boldsymbol{M_{dr}}
    \\
     -\boldsymbol{M_{dr}} \boldsymbol{M_r}^{-1}  & \boldsymbol{J_r} - \boldsymbol{\tilde{F}_r} \boldsymbol{K_{vr}} \boldsymbol{\tilde{F}_r}^{\top}
    \end{array}\right]
    \left[\begin{array}{c}
    {\frac{\partial H_{dr}}{\partial \boldsymbol{r}}}(\boldsymbol{r},\boldsymbol{p_r}) \\
    {\frac{\partial H_{dr}}{\partial \boldsymbol{p_r}}}(\boldsymbol{r},\boldsymbol{p_r})
    \end{array}\right]
\end{equation}
\normalsize
\begin{equation}\label{eq:ham_desired_r_out}
    \boldsymbol{y_{dr}} 
    =
    \boldsymbol{\tilde{F}_r}^{\top} \frac{\partial H_{ds}}{\partial \boldsymbol{p_r}}
    \end{equation}
\begin{equation}\label{eq:ham_desired_r_ham}
    H_{dr}
    =
    \frac{1}{2} \boldsymbol{p_r}^{\top} \boldsymbol{M_{dr}}^{-1}(\boldsymbol{r}) \boldsymbol{p_r}+V_{dr}(\boldsymbol{r})
\end{equation}
 The two systems \eqref{eq:ham_desired_s_eom}-\eqref{eq:ham_desired_s_ham} and \eqref{eq:ham_desired_r_eom}-\eqref{eq:ham_desired_r_ham} are decoupled and each one evolves on its own configuration manifold $\mathcal{S}$ and $\mathcal{R}$, respectively, though with a coupling due to the nonholonomic constraint. Note also that system \eqref{eq:ham_desired_r_eom}-\eqref{eq:ham_desired_r_ham} is of the original unconstrained Hamiltonian form (holonomic). Thus, the unconstrained variables $\boldsymbol{r}$ are easy to stabilize with a smooth control law, whereas for $\boldsymbol{s}$, stabilization is not straightforward. Analytically deriving the energy evolution of \eqref{eq:ham_desired_eom}-\eqref{eq:ham_desired_ham} bearing in mind the decomposition in the previous section yields:
\begin{equation}\label{eq:hddot_evolution}
    \dot{H}_d = \underbrace{\frac{\partial ^{\top} H_d}{\partial \boldsymbol{s}} \boldsymbol{\dot{s}}+\frac{\partial ^{\top} H_d}{\partial \boldsymbol{\tilde{p}_s}}  \boldsymbol{\dot{\tilde{p}}_s}}_{\dot{H}_{ds}} + 
    \underbrace{\frac{\partial ^{\top} H_d}{\partial \boldsymbol{r}} \boldsymbol{\dot{r}}+
    \frac{\partial ^{\top} H_d}{\partial \boldsymbol{p_r}} \boldsymbol{\dot{p}_r}}_{\dot{H}_{dr}}
\end{equation}
With straightforward calculations we can deduct that for the system evolving on $\mathcal{S}$ we have $\dot{H}_{ds} \leq 0$ with: 
\begin{equation}\label{eq:hdsdot=0}
     \dot{H}_{ds} = 0 \Rightarrow
         \mathcal{D}_s^{\top}(\boldsymbol{r})\frac{\partial V_{ds}(\boldsymbol{s})}{\partial \boldsymbol{s}} = \boldsymbol{0}_{n-p-k}
\end{equation}
and for the system evolving on $\mathcal{R}$ we have in a similar manner $\dot{H}_{dr} \leq 0$ with:
\begin{equation}\label{eq:hdrdot=0}
     \dot{H}_{dr} = 0 \Rightarrow 
     \frac{\partial V_{dr}(\boldsymbol{r})}{\partial \boldsymbol{r}} = \boldsymbol{0}_p
\end{equation}
With closer attention to \eqref{eq:hdsdot=0}, we can observe that a promising attempt for stabilization of the constrained variables $\boldsymbol{s}$ is the following: Drive the  $s$-dynamics to the invariant set $\mathbf{\Omega}_{inv}$ at an arbitrary stabilization point, denoted by $\boldsymbol{s^{\omega}}$, while recruiting the $r$-dynamics to "guide" the system from $\boldsymbol{s^{\omega}}$ towards $\boldsymbol{s^*}$ via matrix $\mathcal{D}_s(\boldsymbol{r})$. Using PCD as in \citet{lee2016}, we can proceed with designing a passivity-based switching control law that can asymptotically stabilize the system in any configuration.

\subsection{Proposed Desired Potential for Full-State Stabilization}
\begin{figure}
\begin{center}
\includegraphics[width=7.8cm]{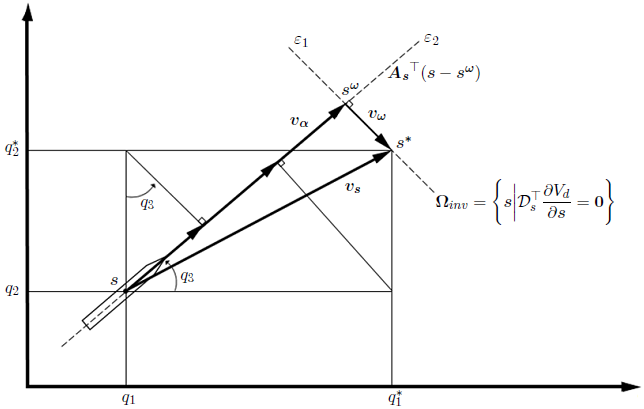}    
\caption{Two independent control actions for the $r$-dynamics and the $s$-dynamics.}
\label{fig::knife_dw2}
\end{center}
\end{figure}

The goal to stabilize the system at a desired configuration $\boldsymbol{q} \rightarrow \boldsymbol{q^*}$ can be achieved sequentially. First, driving $\boldsymbol{s} \rightarrow \boldsymbol{s^*}$ by utilizing the $r$-dynamics and then $\boldsymbol{r} \rightarrow \boldsymbol{r^*}$ with a smooth potential. Thus, according to equations \eqref{eq:hdsdot=0} and \eqref{eq:hdrdot=0} we need to design the desired potential functions $ V_{ds}$ and $ V_{dr}$ for each stabilization task, respectively. Based on \citet{lee2016}, the desired potential $V_{ds}:\ \mathcal{S} \rightarrow \mathbb{R}$ is required to fulfill the following:
\vspace{-1mm}
\begin{enumerate}
    \item $V_{ds} \geq 0$ with the equality holding when $\boldsymbol{s}=\boldsymbol{s^*}$
    \item $\frac{\partial V_{ds}}{\partial \boldsymbol{s}} = 0,\ \text{iff}\ \boldsymbol{s}=\boldsymbol{s^*}$
    \item  $V_{ds}$ is radially unbounded
\end{enumerate}
We begin with the $s$-dynamics for which we choose a quadratic function that satisfies the aforementioned requirements defined as:
\begin{equation}\label{eq:Vds}
    V_{ds} = \frac{1}{2}(\boldsymbol{s} - \boldsymbol{s^*})^{\top}
    \boldsymbol{Q_s}
    (\boldsymbol{s} - \boldsymbol{s^*})
\end{equation}
with $\boldsymbol{Q_s} \in \mathbb{R}^{(n-p) \times (n-p)}$ a constant symmetric matrix serving tuning purposes. With this choice, equation \eqref{eq:hdsdot=0} yields:
\begin{equation}\label{eq:affine_invariant_set} 
    \mathcal{D}_s^{\top}(\boldsymbol{r})\boldsymbol{Q_s}(\boldsymbol{s} - \boldsymbol{s^*}) = \boldsymbol{0}_{n-p-k}
\end{equation} 
which describes a $k$-dimensional affine hyperplane in $\mathcal{S} \in \mathbb{R}^{n-p}$ that is defined by a set of $n-p-k$ linear equations. Thus the system will not be stabilized at $\boldsymbol{s^*}$ but rather at another point denoted by $\boldsymbol{s^{\omega}} \in \mathbf{\Omega}_{inv}$. Let $\boldsymbol{v_s} = \boldsymbol{s} - \boldsymbol{s^*}$ be the vector that we want to drive to zero. Then, matrix $\mathcal{D}_s(\boldsymbol{r})\boldsymbol{Q_s}$ maps this vector to the constrained space as a new vector:
\begin{equation}\label{eq:valpha}
    \boldsymbol{v_{\alpha}} = \mathcal{D}_s^{\top}(\boldsymbol{r})\boldsymbol{Q_s}\boldsymbol{v_s} \in \mathbb{R}^{n-p-k} 
\end{equation}
which is the vector on the constraint space that we are able to drive to zero ($\boldsymbol{v_{\alpha}} \rightarrow \boldsymbol{0}$) with the potential function $V_{ds}$ chosen as in \eqref{eq:Vds}, and acting only on the constrained variables $\boldsymbol{s}$, thus driving $\boldsymbol{s} \rightarrow \boldsymbol{s^{\omega}}$. We continue with the following critical observation: Since we have assumed that the nonholonomic constraints are a function of only $\boldsymbol{r} \in \mathcal{R}$ and act only on $\boldsymbol{s} \in \mathcal{S}$, the constraint equation \eqref{eq:constraint_on_s} is now integrable in $\mathcal{S}$ and can get the form:
\begin{equation}\label{eq:constraint_space}
     \boldsymbol{A_s}^{\top}(\boldsymbol{r})\left(\boldsymbol{s^{\omega}} - \boldsymbol{s} \right) = \boldsymbol{0}
\end{equation}
which describes an $(n-p-k)$-dimensional, affine hyperplane in $\mathcal{S}$, defined by a set of $k$ linear equations. This affine hyperplane describes the constrained space on which the system will evolve on, a subspace of $\mathcal{S}$. We observe here that matrix $\mathcal{D}_s(\boldsymbol{r})\boldsymbol{Q_s}$ maps $\boldsymbol{v_s}$ to the constrained space described in \eqref{eq:constraint_space} and similarly matrix $\boldsymbol{A_s}^{\top}(\boldsymbol{r})$ maps $\boldsymbol{v_s}$ to the invariant set defined in \eqref{eq:affine_invariant_set}:
\begin{equation}\label{eq:vomega}
    \boldsymbol{v_{\omega}} = \boldsymbol{A_s}^{\top}(\boldsymbol{r})\boldsymbol{v_s}  \in \mathbb{R}^{k} 
\end{equation}
Since these spaces are the orthogonal complement of each other, we know that $\boldsymbol{v_{s}} \rightarrow \boldsymbol{0}$ if both $\boldsymbol{v_{\alpha}} \rightarrow \boldsymbol{0}$ and $\boldsymbol{v_{\omega}} \rightarrow \boldsymbol{0}$ is achieved. We have already showed that $\boldsymbol{v_{\alpha}} \rightarrow \boldsymbol{0}$ is feasible for the quadratic choice of $V_{ds}$ in \eqref{eq:Vds}. The concept now is to use the unconstrained variable $\boldsymbol{r}$ in order to drive $\boldsymbol{v_{\omega}}$ to zero as well, meaning that $\boldsymbol{s^{\omega}} \rightarrow \boldsymbol{s^*}$ and thus $\boldsymbol{s} \rightarrow \boldsymbol{s^*}$. This is possible by the following quadratic choice:
\begin{equation}\label{eq:vdr}
    V_{dr} = \frac{1}{2} \boldsymbol{v_{\omega}}^{\top} \boldsymbol{Q_r} \boldsymbol{v_{\omega}}
\end{equation}
with $\boldsymbol{Q_r} \in \mathbb{R}^{k \times k}$ a constant symmetric matrix for tuning purposes. Thus the system will be stabilized at $\boldsymbol{v_{\omega}} = \boldsymbol{0}$, and since $\boldsymbol{v_{\alpha}} = \boldsymbol{0}$ can be driven to zero we conclude that we obtain $\boldsymbol{v_{s}} = \boldsymbol{0}$ implying $\boldsymbol{s} \rightarrow \boldsymbol{s^*}$. Note that $\boldsymbol{v_{\alpha}} = \boldsymbol{v_{\alpha}}(\boldsymbol{s},\boldsymbol{r})$, $\boldsymbol{v_{\omega}} = \boldsymbol{v_{\omega}}(\boldsymbol{s},\boldsymbol{r})$, meaning that both the desired potentials are functions of both the constrained and unconstrained variables implying a coupling of the systems via the potential components of the control law \eqref{eq:feedback_law}. However, due to orthogonality, each desired potential $V_{ds}$ and $V_{dr}$ leads to potential forces that act only on their respective variables. More specifically, the control action on $\mathcal{S}$ is the term $\mathcal{D}_s^{\top}(\boldsymbol{r})\boldsymbol{Q_s}\boldsymbol{v_s} \in \mathbb{R}^{n-p-k} $. While it is a function of $\boldsymbol{r}$, it only acts on the $\boldsymbol{s}$ variables. Similarly, the control action on $\mathcal{R}$ is given by $\frac{\partial V_{dr}}{\partial \boldsymbol{r}} =\frac{\partial^{\top} \boldsymbol{v_\omega}}{\partial \boldsymbol{r}} \boldsymbol{Q_r} \boldsymbol{v_\omega}$. While it is a function of $\boldsymbol{s}$, it acts only on $\boldsymbol{r}$. The aforementioned observations are illustrated graphically for the simple knife-edge example in Figure \ref{fig::knife_dw2}. 
Having achieved $\boldsymbol{s} \rightarrow \boldsymbol{s^*}$ (i.e., stabilizing the constrained variables $\boldsymbol{s}$ which are in general difficult to handle), we can shift our attention to the unconstrained variables $\boldsymbol{r}$. The unconstrained variables $\boldsymbol{r}$ are not stabilized on the desired equilibrium $\boldsymbol{r^*}$ since they were used so far to stabilize the other variables. Now that $\boldsymbol{s} = \boldsymbol{s^*}$ we can switch to another simple quadratic desired potential function for $\boldsymbol{r}$ and since these variables are not hindered by constraints they can be stabilized to the desired equilibrium $\boldsymbol{r^*}$. Note that the aforementioned control choices lead to asymptotic stabilization, which is more of a theoretical interest as $\boldsymbol{s} \rightarrow \boldsymbol{s^*}$ converges over infinitely long time. For this reason, we can attain $\boldsymbol{r} \rightarrow \boldsymbol{r^*}$ in practice by triggering the switch when the norms $\norm{\boldsymbol{s}-\boldsymbol{s^*}}$ and $\norm{\boldsymbol{\dot{\tilde{s}}}}$ are small enough (i.e., setting stopping criteria $s^d$ and $\dot{s}^d$ respectively). Moreover, setting the ground for collision avoidance in the constrained space, we can implement other expressions for the desired potential $V_{ds}$ that incorporate repulsive fields for the purposes of collision avoidance according to \citet{khatib1985}. Thus, we can express the control action for the unconstrained variables more generally as:
\begin{equation}\label{eq:vomegavds}
    \boldsymbol{v_{\omega}} = \boldsymbol{A_s}^{\top}(\boldsymbol{r})\frac{\partial V_{ds}(\boldsymbol{s})}{\partial \boldsymbol{s}}
\end{equation}
and define the desired potential to stabilize $\boldsymbol{r}$ as: 
\begin{equation}\label{eq:V_dr_branch}
    V_{dr} =
\left\{
    \arraycolsep=1.4pt\def\arraystretch{2.2}
	\begin{array}{ll}
	\frac{1}{2} \frac{\partial^{\top} V_{ds}(\boldsymbol{s})}{\partial \boldsymbol{s}}  \boldsymbol{A_s} \boldsymbol{Q_r} \boldsymbol{A_s}^{\top} \frac{\partial V_{ds}(\boldsymbol{s})}{\partial \boldsymbol{s}} \\
	\frac{1}{2}(\boldsymbol{r} - \boldsymbol{r^*})^{\top}
    \boldsymbol{Q_r}(\boldsymbol{r} - \boldsymbol{r^*}), \; \mbox{if } \norm{\boldsymbol{s}-\boldsymbol{s^*}} < s^d
	\end{array}
\right.
\end{equation}
where $V_{ds}$ can take the form \eqref{eq:Vds} or a more general one that satisfies the assumptions made earlier in this section. Note that in case the nonholonomic system is underactuated, the desired potentials chosen as in \eqref{eq:Vds} and \eqref{eq:V_dr_branch} need to satisfy the matching conditions \eqref{eq:mathcing_cond_kin} and \eqref{eq:mathcing_cond_pot}.
\begin{thm}   
Consider a nonholonomic system described by \eqref{eq:ham_exp_constr_eom}-\eqref{eq:ham_exp_constr_ham}, satisfying the assumptions made in Subsection \ref{ss:PCD_ham}. Control law \eqref{eq:feedback_law} satisfying the matching conditions \eqref{eq:mathcing_cond_kin}-\eqref{eq:mathcing_cond_pot}, and desired potentials chosen as in \eqref{eq:Vds} and \eqref{eq:V_dr_branch}, will stabilize the system to the desired equilibrium \eqref{eq:goal_qstar}.
\end{thm}
\begin{pf}    
As explained in this section, such a system can be decomposed to two independent ones. For each system we can see the evolution of the storage function. For the $s$-dynamics we have ${H}_{ds} > 0$, $\dot{H}_{ds} \leq 0$ with $\dot{H}_{ds} = 0$ iff, $\mathcal{D}_s^{\top}(\boldsymbol{r})\frac{\partial V_{ds}(\boldsymbol{s})}{\partial \boldsymbol{s}} = \boldsymbol{0}_{n-p-k}$ meaning that the system will converge to this equilibrium. Similarly, for the $r$-dynamics we have ${H}_{dr} > 0$, $\dot{H}_{dr} \leq 0$ with $\dot{H}_{dr} = 0$ iff, $\frac{\partial V_{dr}(\boldsymbol{r})}{\partial \boldsymbol{r}} = \boldsymbol{0}_p$. Then, selecting $V_{dr}$ as in the first branch of \eqref{eq:V_dr_branch}, $\frac{\partial V_{dr}(\boldsymbol{r})}{\partial \boldsymbol{r}} = \boldsymbol{0}_p \Rightarrow \frac{\partial V_{ds}(\boldsymbol{s})}{\partial \boldsymbol{s}} = \boldsymbol{0}_{n-p-k}$ thus $\boldsymbol{s} \rightarrow \boldsymbol{s^*}$ and stabilization in the constrained space is achieved. Switching $V_{dr}$ as in the second branch of \eqref{eq:V_dr_branch} will yield $\boldsymbol{r} \rightarrow \boldsymbol{r^*}$ and as such, full-state stabilization is achieved $\boldsymbol{q} \rightarrow \boldsymbol{q^*}$.
\end{pf}

\section{Simulation Results}\label{ss:results}
In this section, we show the efficacy of the aforementioned results in two different simulations scenarios. For brevity, we do not present a detailed description of the approach with which the single-agent results are extended in a distributed setting. The original approach of how to extend IDA-PBC in a distributed manner can be found in \citet{valk2018}. In \citet{tsolakis2021}, a step-by-step adaptation of distributed IDA-PBC to nonholonomic systems is described along with the implementation of a simple collision avoidance protocol that relies on the APF method.

\begin{figure}
\begin{center}
\includegraphics[width=7.8cm]{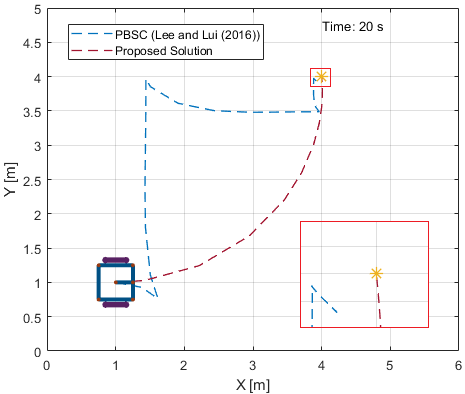}    
\caption{Trajectory comparison between the solution proposed here and PBSC proposed in \citet{lee2016}.}
\label{fig::sc1_traj}
\end{center}
\end{figure}

\begin{figure}
\begin{center}
\includegraphics[width=7.8cm]{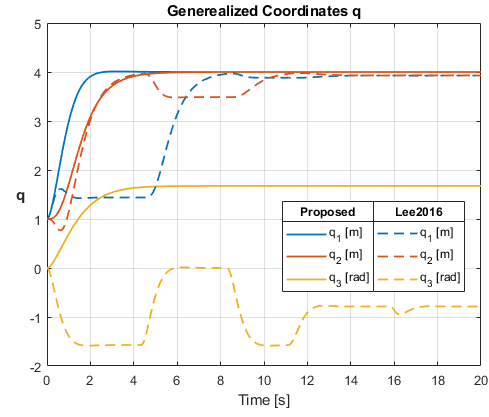}    
\caption{The generalized coordinates of the differential robot for each approach where faster convergence with the proposed solution is clear.}
\label{fig::sc1_q}
\end{center}
\end{figure}

\begin{figure}
\begin{center}
\includegraphics[width=7.8cm]{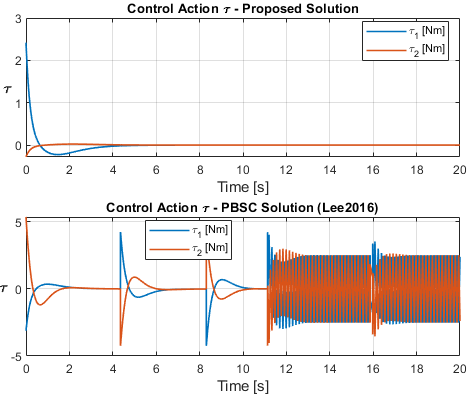}    
\caption{Comparison of the control actions. With the proposed solution smooth stabilization can be achieved in the constrained space.}
\label{fig::sc1_tau}
\end{center}
\end{figure}

We first show a comparison for a single-agent scenario featuring a differential robot. The differentially driven wheeled robot starts from initial configuration $\boldsymbol{q_0} = \begin{pmatrix} 1 & 1 & 0 \end{pmatrix}^\top$ and has the goal configuration $\boldsymbol{q^*} = \begin{pmatrix} 4 & 4 & \theta \end{pmatrix}^\top$ (with $\theta$ denoting a free orientation). We compare two different trajectories: First, the trajectory that our solution yields after applying \eqref{eq:feedback_law} with desired potential functions as in \eqref{eq:Vds} and the first branch of \eqref{eq:V_dr_branch}. Then, the trajectory that the Passivity Based Switching Control (PBSC) from \citet{lee2016} yields. We can see in Figure \ref{fig::sc1_traj} that with the approach proposed here we can achieve smooth stabilization in the constrained space. This results in faster convergence to the goal as observed in Figure \ref{fig::sc1_q} and also no oscillatory phenomena in the control action which can be seen for the PCSC solution in Figure \ref{fig::sc1_tau}. Full-state stabilization can be achieved after switching to the second branch of \eqref{eq:V_dr_branch}.

\begin{figure}
\begin{center}
\includegraphics[width=7.8cm]{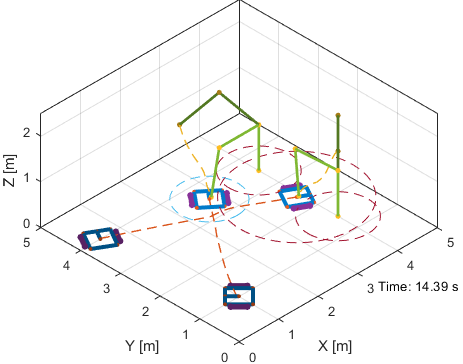}    
\caption{Trajectories in the multi-agent scenario of two differential robots and two manipulators.}
\label{fig::sc4_traj}
\end{center}
\end{figure}

\begin{figure}
\begin{center}
\includegraphics[width=7.8cm]{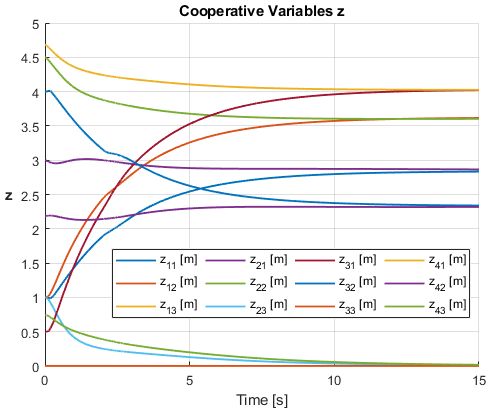}    
\caption{Consensus of the cooperative variables $\boldsymbol{z}$ of the two differential robots and two manipulators.}
\label{fig::sc4_z}
\end{center}
\end{figure}

We then illustrate a multi-agent scenario in which we have two differential robots and two 3-DoF manipulators. The goal is for each pair of differential robot and manipulator to reach consensus emulating a practical application in which for example we want to unload cargo from the differentially driven robots. We see in Figure \ref{fig::sc4_traj} the trajectories of these robots and in Figure \ref{fig::sc4_z} that consensus is being reached for all cooperative variables. These cooperative variables $\boldsymbol{z}$ are the $xyz$-coordinates of each agent. This is an example of how the distributed extension can work for heterogeneous agents. Note that in place of the manipulators, underactuated systems such as quad-rotors or overhead cranes can be used for which IDA-PBC solutions already exist. Animations of the simulation examples can be found in \cite{vidplaylist}.

\section{Conclusions}\label{ss:conclusion}
In this work, we have extended a recently proposed distributed control method from \cite{valk2018} to the widely applicable class of nonholonomic mechanical systems. This framework allows the design of a unified, distributed control law for a team of under-actuated and/or nonholonomic heterogeneous mechanical systems. For future work we aim to broaden the scope of application by investigating how the proposed method can work for other nonholonomic agents and particularly for agents that are also underactuated (e.g., the inverted mobile pendulum).


\bibliography{ifacconf}             

\end{document}